\documentclass[letterpaper]{article}

\usepackage{natbib,alifeconf}  
\usepackage{amsmath}
\usepackage{url,hyperref,cleveref}
\usepackage{booktabs}
\usepackage{graphicx}
\usepackage{siunitx}
\usepackage{tabularx}
\usepackage{hyperref}
\usepackage{nameref}
\title{Towards Robust Universal Perturbation Attacks: A Float-Coded, Penalty-Driven Evolutionary Approach}

\author{Shiqi (Edmond) Wang$^{1}$, Mahdi Khosravy$^{2}$, Neeraj Gupta$^3$, \and Olaf Witkowski$^2$\\    \mbox{}\\    
$^1$ University of California, Los Angeles, California, USA\\    
$^2$ Cross Labs, Cross-Compass Ltd., Tokyo, Japan\\    
$^3$ Oakland University, Rochester, MI, USA\\   
} 

\begin{document}

\maketitle

\begin{abstract}
    Universal adversarial perturbations (UAPs) have garnered significant attention due to their ability to undermine deep neural networks across multiple inputs using a single noise pattern. Evolutionary algorithms offer a promising approach to generating such perturbations due to their ability to navigate non-convex, gradient-free landscapes. In this work, we introduce a float-coded, penalty-driven single-objective evolutionary framework for UAP generation that achieves lower visibility perturbations while enhancing attack success rates. Our approach leverages continuous gene representations aligned with contemporary deep learning scales, incorporates dynamic evolutionary operators with adaptive scheduling, and utilizes a modular PyTorch implementation for seamless integration with modern architectures. Additionally, we ensure the universality of the generated perturbations by testing across diverse models and by periodically switching batches to prevent overfitting. Experimental results on the ImageNet dataset demonstrate that our framework consistently produces perturbations with smaller norms, higher misclassification effectiveness, and faster convergence compared to existing evolutionary-based methods. These findings highlight the robustness and scalability of our approach for universal adversarial attacks across various deep learning architectures. 
\end{abstract}

\section{Introduction}

With deep learning becoming the cornerstone of modern machine learning, powering applications ranging from image classification to natural language understanding and autonomous control \cite{deep_learning_paper}, \cite{ImageNet}. Their success can be attributed to deep learning network's ability to learn complex feature hierarchies and form accurate predictions. Nevertheless, as these models become ubiquitous in safety and security-critical environments, concerns about robustness and trustworthiness have grown. Specifically, adversarial vulnerabilities - small and often imperceptible perturbations - can cause significant misclassifications which threaten the reliability of neural networks \cite{deep_learning_book}. These attacks span a broad category of adversarial attacks capable of degrading or even subverting network performances \cite{adversarial_examples}, \cite{model_inversion}.

Among these attacks, Universal Adversarial Perturbations (UAPs) stand out for their ability to consistently mislead a model across multiple inputs \cite{uap}. Rather than crafting a unique perturbation for each image, UAPs aim to discover a single noise pattern that degrades performance on a large subset of the dataset. This property poses an urgent security risk because it drastically reduces the attacker's overhead: the same configuration can be applied almost "universally", undermining even high-performing networks from convolutional neural networks (CNNs) to vision transformers (ViTs) \cite{resistant_models,image_retrieval,adversarial_images,intriguing_properties}. Traditionally, generating such universal noise leverages optimization-based approaches \cite{crafting_uap}, \cite{nag}, or generative models \cite{gen_mod}. Beyond image classification, universal perturbation attacks have proliferated into text classification \cite{text_uap}, image retrieval \cite{image_retrieval}, segmentation \cite{data_free}, and object detection \cite{obj_det}. Yet, evolutionary methods, especially Genetic Algorithms (GA), have seldom been explored for universal attacks.

In an earlier attempt to fill this gap, the Evolutionary Universal Perturbation Attack (EUPA) \cite{eupa} introduced a nonlinear evolutionary search for effective universal noise. EUPA employed a genetic algorithm to craft a perturbation that minimizes visibility while maximizing classification. They explored two optimization strategies - \textbf{Constrained single-objective EUPA} (SC-EUPA), enforcing a noise threshold (visibility constraint), \textbf{and Pareto double-objective EUPA} (PD-EUPA), jointly optimizing for minimal noise and maximal attack intensity without a fixed visibility constraint. The paper asserted SC-EUPA to be superior in attack generation, since the minimization of visibility is somewhat unnecessary as long as it is below a fixed constraint. 

\subsection{Our Contributions}

In this work, we build on \cite{eupa} and advance a modern, robust float-coded framework for generating EUPAs:

\begin{itemize}
    \item \textbf{Floating-Point Representations}:
    Instead of integer-coded perturbations, we adopt continuous genes to capture gradient-like changes in a smoother space. This aligns with current deep learning scales and yields subtler, more adaptive perturbations.
    \item \textbf{Penalty-driven Single Objective Optimization}:
    We operate within a $\epsilon$-constraint framework, enforcing a norm limit while maximizing misclassification. We deviate from a threshold-based method to offer a flexible and more efficient way to handle the norm-accuracy trade-off. We also disable elitism for enhanced norm reduction, requiring solutions to always re-qualify and preventing the same solution from always staying on top.
    \item \textbf{Dynamic Operators}:
    We do not rely on static probabilities or integer-coded flips. We implement a linear decay scheduling for the evolutionary operators of crossover \& mutation, and an exponential one for the $\epsilon$ constraint. We also create a conditional pixel-cleaning operator that prevents the algorithm from continually abating the norm even after the constraint is satisfied.
    \item \textbf{Extensive Pytorch Implementation}:
    Our pipeline is also developed with PyTorch and PyGAD \cite{pygad} in a modular structure rather than an integer-coded MATLAB setup. This simplifies integration with standard deep learning frameworks and allows easy extension to other architectures.
    \item \textbf{Extended Universality}:
    We also prevent overfitting by switching the image batches every $k$ generations, as well as testing the perturbation on three diverse and widely used computer vision models. This ensures the universal perturbation is truly universal in both data and network architectures.
    
\end{itemize}

We show that our approach consistently produces lower final perturbation norms while achieving greater misclassification effectiveness. It converges significantly faster in moderate population sizes and remains straightforward to extend to additional models. 

\subsection{Scope and Outline} 

In \textit{Methodology}, we formalize our single-objective, penalty-based universal adversarial perturbation framework, detailing how a GA can target both low norm and high misclassification without optimizing for both simutaneously. \textit{Experiments} discusses the specific architecture of our evolutionary search - covering float-coded chromosomes, pixel cleaning, and dynamic operator scheduling. \textit{Results} presents results demonstrating that the proposed method achieves robust universal attacks on GoogLeNet \cite{googlenet}, Resnet50 \cite{resnet}, and ViT-B/16 \cite{vit} while maintaining minimal visible perturbations. Finally, \textit{Discussion} concludes the paper by summarizing our findings and suggesting future extensions, including more advanced operator designs, multi-model ensemble objectives, and broader application contexts.

\section{Methodology} 

In this section, we briefly describe the mathematical formulation and optimization approaches for designing the Evolutionary Universal Perturbation Attack (EUPA). For complete details, see \cite{eupa}. We focus on the Constrained Single-Objective EUPA (SD-EUPA) which is proven to be more effective than the Pareto Dual-Objective version (PD-EUPA). In particular, we give an overview of the methodologies for modelling the attack, relevant symbols and their relations to traditional evolutionary algorithms, along with the optimization characterization.

\section{1. Attack Formation}  
\label{sec:attack}

\subsection{A. Attack Intensity \& Perturbation Visibility}
We begin with the characterization of the attack intensity. We measure this by Expected Misclassification $\Gamma$ (we also refer to this as the attack intensity), which quantifies the extent our perturbation affects the model classification performance. The attack aims to maximize the misclassification:

\begin{equation}
    \Gamma = \sum^M_{m=1} \alpha_m p(\alpha_m)
\end{equation}

where $\alpha_m$ is the rate of misclassification for the $m$-th data point, and $p(\alpha_m)$ is the probability of a given misclassification. We have $M$ data points in total. We want to maximize the ability for the perturbation to cause misclassification, changing the form $f$ (vanilla output) to $\hat{f}$ (output after perturbation). 

To minimize the perceptibility of the perturbation, we also measure the perturbation visibility using Mean Squared Error (MSE):

\begin{equation}
    \xi = \frac{1}{3\,n_{\text{pix}}} \sum_{c=1}^{3} \sum_{k=1}^{n_{\text{pix}}} \bigl(P_k - \hat{P_k}\bigr)^2
\end{equation}

where $P_k$ and $\hat{P_k}$ represent the pixel value of the original and perturbed images, $n_{pix}$ is the number of pixels per color channel. We calculate the MSE over all pixels and over the three RGB color channels, averaged over $3p$ of them.

\subsection{B. Pareto Double-Objective Optimization}
For the Pareto Double-Objective Optimization (PDO), we have dual conflicting objectives to simultaneously maximize attack intensity and minimize perturbation visibility. 

The problem can be formulated as:

\begin{align}
    \max \theta_1 \min \theta_2 
    \begin{cases}
         \theta_1 = \Gamma  \\
         \theta_2 = \xi
    \end{cases}
\end{align}

Here $\theta_1$ and $\theta_2$ are the same as before, representing attack intensity and perturbation visibility, respectively. We solve this PDO with a population-based evolutionary algorithm in order to find the best balance between maximizing misclassification and minimizing perceptibility.

\subsection{C. Constrained Single-Objective Optimization}
For the Constrained Single-Objective problem (CSO), we focus on maximizing attack intensity and place a simple constraint for the perturbation visibility.

The problem can be formulated as:

\begin{align}
    \max \theta &= \Gamma \\
    s.t. \Delta_k &\in [\Delta^l_k, \Delta^u_k]
\end{align}

where $\Delta_k$ (perturbation at the $k$-th pixel is constrained by the lower and upper bounds of the perturbation $\Delta^l_k$ and $\Delta^u_k$.

\subsection{D. Single Objective vs Pareto Double-Objective}

Here, both approaches seem reasonable, but the dual-objective problem (PDO) has a problem where the objectives are inherently conflicting: increasing $\Gamma$ may require more noticeable perturbations, thereby increasing $\xi$. In \cite{eupa}, it is proven by empirical results that the PD-EUPA method works more efficiently than the CS-EUPA. In this analytic review, we confirm this result and also provide a mathematical foundation for why this is the case.

First, we introduce the \textbf{Weighted Sum Method}, which transforms a multi-objective problem into a single-objective one. It is very straightforward and proceeds by assigning weights to each objective and summing:

\begin{equation*}
    \textbf{Maximize} \  F = w_1\Gamma - w_2\xi
\end{equation*}

where $w_1$ and $w_2$ are positive weights reflecting the relative importance of each objective. Nevertheless, this method is often limited in practice. First, the weighted sum method is typically effective only for problems with convex Pareto fronts. Non-convex regions may not be adequately explored. Furthermore, we note that selecting appropriate weights $w_1$ and $w_2$ is non-trivial and requires extensive experimentation. Inappropriate weight selection $(w_1, w_2)$ can lead to suboptimal solutions - especially if the Pareto front is non-convex \cite{weighted_drawbacks}.

Instead, we choose the \textbf{Epsilon-Constraint Method} \cite{epsilon} which states:

\begin{align*}
    &\textbf{Maximize} \ \Gamma \\
    &\textbf{Subject to} \  \xi \leq \epsilon
\end{align*}

Here we have direct control over the constraint and allows exploration of different levels of $\epsilon$. This approach is preferred, and essentially, we can view the CSO problem as an application of the epsilon-constraint method on the PDO. 

By treating $\xi$ as a constraint, the CS-EUPA allows the optimization algorithm to concentrate solely on maximizing $\Gamma$ within the feasibility region defined by $\xi \leq \epsilon$. When we compare the two approaches, the PD-EUPA feasible set spans:

\begin{equation*}
    \mathcal{S}_{\text{PD}} = \{(\Gamma(\Delta), \xi(\Delta)) | \Delta \in \Delta_{all}\}
\end{equation*}

Whereas the CS-EUPA reduces it to:

\begin{equation*}
    \mathcal{S}_{\text{CS}} = \{ \Delta \mid \xi(\Delta) \leq \epsilon \}
\end{equation*}

Compared to the above, we reduce solutions with $\xi(\Delta) > \epsilon$, allowing the GA to explore a more targeted region of the search space. This can yield faster convergence under certain search heuristics \cite{multiobjective}. Analytically, solving $\max\Gamma$ subject to $\xi\le\epsilon$ can also be simpler in some problem formulations \cite{nonlinear_multiobjective}. The Lagrangian form can provide direct insight into how solutions evolve with respect to $\epsilon$. 


\subsection{E. Perturbation Bounds}
The criterion for selecting perturbation bounds is also interesting. We constrain each pixel of the universal perturbation $\Delta$ by empirical bounds derived from the training set standard deviation. Specifically, let $\sigma_k$ be the standard deviation of the k-th pixel across all images. We then assign:

\begin{align*}
    &\begin{cases}
        \Delta^l_k = -\sigma_k \\
        \Delta^u_k = \sigma_k \\
    \end{cases}
    \Rightarrow \Delta_k \in [-\sigma_k, \sigma_k] \\\\
    &\text{where} \   \sigma^c_k \;=\; \sqrt{\frac{1}{m}\sum_{j=1}^{m}\bigl(\Lambda_{j,k}^c - \phi_{k}^c\bigr)^2}
\end{align*}

Where $\phi^c_k$ is the mean pixel value over all images at that pixel value for that color. 

Empirically, we found that this ensures the perturbation remains relatively imperceptible while maximizing $\Gamma$.

\section{Genetic Algorithms Framework for UAP } \label{sec:ga}
 For the task of generating universal adversarial perturbations, we employ a Float-Coded Genetic Algorithm (FCGA) within a Penalty-Drive Single-Objective framework. 
GA begins when we initialize a population of chromosomes, each representing a candidate universal perturbation $\Delta$. The population $P$ is defined as:
\begin{equation}
P = \{\gamma_i\}^r_{i=1}
\end{equation}
where $r$ denotes population size. 
Each chromosome $\gamma_i$ is a 3D Tensor (height x width x RGB):
\begin{equation}
\gamma_i = \{g^i_{k, c}\} \ \text{for} \    c \in \{1, 2, 3\}, k \in \{1, \dots, n_{pix}\}
\end{equation}
Here $g^i_{k,c}$ represents the perturbation value applied to the $k$-th pixel in the $c$-th color channel. To start with, we set most pixel values to $0$, with a fraction sampled in $[\Delta^l_k, \Delta^u_l]$ for low visibility. 
Encapsulated by $\gamma_i$, the universal perturbation $\Delta$ representing the differences between original and perturbed images: 
\begin{equation}
[\Delta = \hat{\Lambda_j} - \Lambda_j \quad \forall j\in \{1, \dots, M\}
\end{equation}
Our encoding strategy utilizes continuous values for perturbation genes, facilitating smooth, gradient-like modifications through genetic operations such as crossover and mutation. 

As specified in the \ref{sec:attack} section, the fitness function is parimarily $\Gamma$, with $\xi \leq \epsilon$ remaining as a constraint:
\begin{align*}
    \Gamma(\gamma_i) &= \sum^M_{m=1}\alpha_mp(\alpha_m) \\
    \xi(\gamma_i) &= \frac{1}{3n_{\text{pix}}} \sum_{c=1}^{3} \sum_{k=1}^{n_{\text{pix}}} (P_k - \hat{P}_k)^2
\end{align*}
Here $P_k$ and $\hat{P_k}$ denote the original and perturbed pixel values. 

Then, the selection mechanism is the operator responsible for choosing parent chromosomes that will undergo crossover to produce offspring. We utilize the \textbf{\textit{Tournament Selection}} method \cite{ga_in_ml}, a robust method in evolutionary computation for balancing exploration and exploitation. 

Specifically, a subset of chromosomes is randomly chosen from the population, and the chromosome with the highest fitness within the subset becomes a parent. The tournament size is a parameter - a larger size increases the likelihood of selecting fitter individuals and accelerating convergence, but also risks convergence, which is premature and not fully representative \cite{intro_to_ga}.

Thereafter, the Crossover operator produces offspring by combining genetic information from two parent chromosomes. We implement a \textbf{\textit{Uniform Crossover}} mechanism - each gene in the offspring is independently chosen from either parent with a probability of 0.5 \cite{artificial_systems}. 

Each gene has the same chance of inheriting characteristics from either parent to maintain diversity.




At this stage, the mutation introduces random alterations and prevents the GA from becoming trapped in local optima. Our mutation strategy involves \textbf{gene re-randomization} \cite{ga}, where a fraction $p_m$ of chromosomes undergo mutation, and within each selected chromosome, each gene has a small probability $p_{\text{flip}}$ of being replaced with a new random value sampled uniformly from the predefined range:

\begin{equation}\nonumber
g_{k,c}^{o} \leftarrow
\begin{cases}
    \text{Random value within } [\Delta_k^l, \Delta_k^u] & \text{with probability } p_{\text{flip}} \\
    g_{k,c}^{o} & \text{otherwise}
\end{cases}
\end{equation}



In contrast to traditional GA approaches that employ elitism to preserve the best solutions, we disable it for our framework. This design choice ensures that newly generated offspring - often with smaller perturbation norms - have the opportunity to surpass older high-fitness yet large-norm solutions. This is to some extent necessary due to our fitness function, where $\xi$ is dynamic and changes every generation.


Each evolutionary epoch consists of \textit{selection}, \textit{crossover}, \textit{mutation}, and \textit{elite selection}. This continues until one of the following termination criteria is met: (i) Maximum number of generations $I_{max}$, (ii) Convergence Threshold, or (iii) Desired fitness.
We can express the above formally as:
\begin{align}
    \text{Terminate} &\ \  \text{if} \ &I \ge I_{max} \\
    &\text{or} \  &|\Gamma_I-\Gamma_{I-1}| < \delta \\
    &\text{or} \ &\Gamma_I \ge \Gamma_{desired}
\end{align}

with $I$ denoting the current generation and $\delta$ for the convergence threshold. 
\subsection{I. Dynamic Operator Scheduling}
To enhance the GA's efficiency and adaptability, we implement dynamic scheduling for the fitness, crossover, and mutation operators:
\begin{itemize}
    \item $p_{\text{crossover}}$: Linearly decays from $0.9\sim0.4$ over the generations.
    \item $p_{\text{mutation}}$: Similarly, linearly decays from $0.6\sim0.2$ over the generations.
    \item $\epsilon$: Exponentially decays from $85\sim35$ over the generations.
\end{itemize}
This scheduling allows the GA to explore broadly in early generations and focus on fine-tuning perturbations as the search progresses.

\begin{figure*}[t]  
\centering
\includegraphics[width=.8\textwidth]{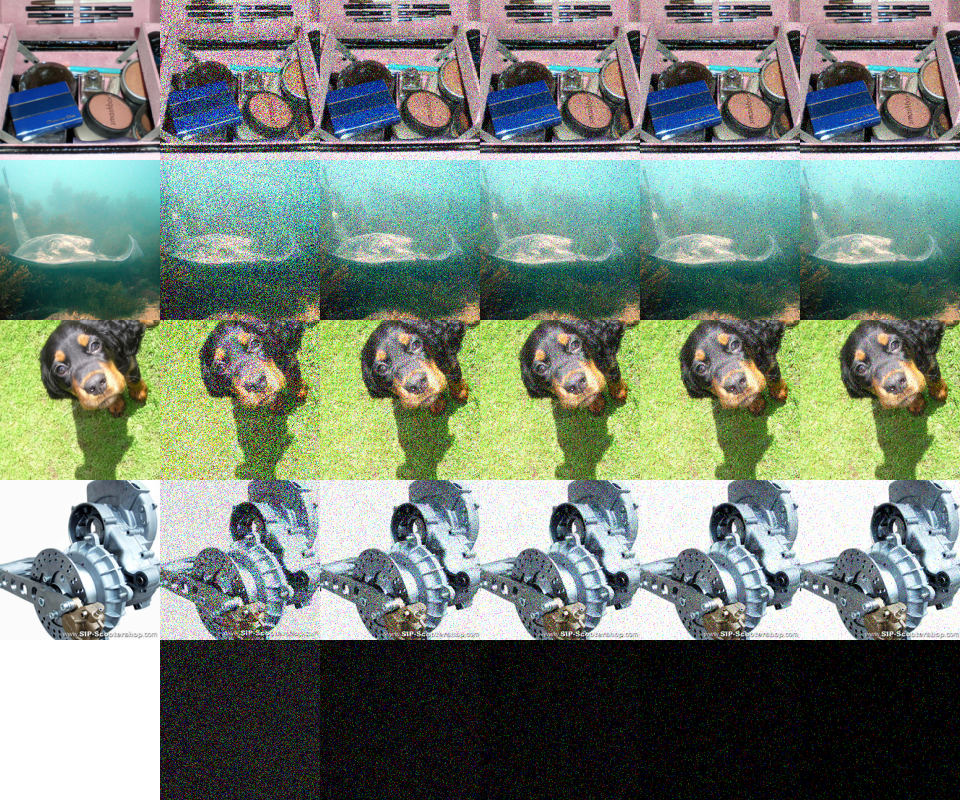}
\caption{Perturbation at generations 1, 16, 32, 48, and 64. The first column shows 4 images taken from the ImageNet dataset, and the second to sixth columns show the same images with perturbations superimposed on top. The curtailment of the perturbation magnitude is very apparent, with the noise almost converging after generation 48. Furthermore, the misclassification score actually increases from 0.5156 to 0.5469 in interactions from gen. 1 to gen. 64; showing an increase in effectiveness while making the perturbation less visible. Just for illustration purposes, we periodically adjust the image batch to prevent overfitting.
}\label{fig:perturbation_start_end}
\end{figure*}
\begin{table*}[h]
\centering
\begin{tabular}{|c|c|c|c|} \hline
\textit{Network} & \textit{\# parameters} & \textit{Accuracy} & \textit{Accuracy (post perturbation)} \\ \hline
GoogLeNet & 6624904 & 0.698 & 0.406 \\ \hline
ResNet-50 & 25557032 & 0.761 & 0.309 \\ \hline
ViT-B/16 & 86567656 & 0.811 & 0.582 \\ \hline
\end{tabular}
\caption{How our perturbation affects the accuracy of three models with different architectures and numbers of parameters. We show their accuracy drops after applying a perturbation acquired after only 64 generations.}
\label{tab:three_networks}
\end{table*}

\section{Experiments} \label{sec:experiments}
In this section, we present a modern PyTorch-compatible EUPA framework built on PyGAD \cite{pygad} with custom processes.
While we draw significant inspiration from the methodology of Gupta et al. \cite{eupa}, we made several practical modifications to the algorithm with varying design motivations. This is primarily to accommodate for floating-point image representations in Python, exploit penalty-based constraints, and adopt dynamic scheduling for certain genetic operators. Below, we outline the key points of improvements and the rationale behind specific hyperparameter choices.
We emphasize how different sets of configurations (penalties, pixel cleaning probabilities, mutation \& crossover rates, etc) affect the perturbation magnitude and misclassification scores.

\subsection{1. Integer Encoding}
Gupta et al. \cite{eupa} define each gene as an integer in a pre-defined range $[\Delta^l_k, \Delta^u_k]$. This ensures a direct manipulation of pixel values and reflects discrete perturbations. This use of an Integer-Coded Genetic Algorithm (ICGA) simplifies constraint handling and mimics discrete pixel modifications. 

Nevertheless, the original implementation was in MatLab which made sense as the images were represented in this way (8-bit grayscale, or 24-bit RGB). For modern PyTorch-based deep learning frameworks, computer vision models expect inputs in a normalized domain. An ImageNet-style normalization typically involves preprocessing the data to have fixed means and standard deviations. With PyTorch and the models expecting float-coded data, a float-coded GA approach is more natural and standard. 

\subsection{2. Fitness Function}
There are a few valid ways to approach the mathematical representation of the CSO problem. Gupta et al. \cite{eupa} chose to bound the perturbation $\Delta$ by fixed thresholds $(\Delta^l_k, \Delta^u_k)$, giving zero reward to all solutions that exceed the bounds. 

First, being based on integer encodings, the rationale behind this approach is somewhat incompatible with modern implementations of deep learning frameworks. In addition, we found an approach bounding every specific pixel in a float-represented setting to be overly restrictive. It also ignores the signal pertaining to the norm of the perturbation $\xi$, a crucial piece of information which could be used to direct the optimization process. 

As a result, we represent a penalty factor $\lambda$ to utilize this signal for compressing norms:
\begin{equation}
    \text{fitness} = \Gamma(\Delta)-\lambda\max(0, \|\Delta\|-\epsilon)
\end{equation}

where $\epsilon$ represents the bound for perturbation visibility $\xi$. Any solution that exceeds $\epsilon$ will see a decrement in net fitness. This more flexible "soft-penalty" system smoothly penalizes large norms instead of cutting off the signal altogether.

Furthermore, we employ an exponential decay schedule for $\epsilon$, starting at a higher threshold and converging to a tighter bound. 

\begin{equation}
    \epsilon_g = \epsilon_{start} \times (\frac{\epsilon_{end}}{\epsilon_{start}})^{\frac{g}{G}}
\end{equation}

where $g$ represents the current generation and $G$ the total number of generations. 

We found a continuous penalty factor simpler to integrate with floating-point representations of images, allowing partial improvements in $\|\Delta\|$ to overshadow purely large perturbations. Furthermore, the exponential decay encourages solutions to slowly reduce norms rather than forcing them under a fixed threshold from the outset. The beginning generations have a stronger incentive to reduce norm to keep up with the steeply diminishing $\epsilon$ constraint, while later generations encourage the solutions to focus on increasing misclassification, with a gentler-declining constraint. We keep the penalty factor $\lambda$ constant as the decaying constraint sufficiently exhibits the dynamic nature we desire. 

\subsection{3. Mutation \& Crossover}
The mutation process of Gupta et al. \cite{eupa} involves an "integer-flip" logic, which we largely stay faithful to. 

In particular, we rely on random reassignments within the range of the initialized genes. We first select a subset of the population $p_m$ to mutate. For each chosen chromosome, a tiny fraction of the genes (we set as $0.005$) become flipped, get reassigned to a new random value in $[\Delta^l_k, \Delta^u_k]$. 

In addition, we explored adding a small float within a pre-defined bound to each mutated gene rather than reassigning them. This seemed too mild and slow to converge. We also explored reassigning a larger percentage of the genes in each chromosome, but this increased aggressiveness often wipes out any successful configuration from previous generations. A subtler dynamic involving sub-populations, in conjunction with the pixel cleaning operation, can often lead to more selective improvements. 

Moreover, we introduced linear decay for both the probabilities of crossover and mutation operators. This can foster broader exploration early on and finer exploitation later, without needing the stark contrast seen in the exponential decay scheduling.

\subsection{4. Pixel Cleaning Operation}
Gupta et al. \cite{eupa} introduced a pixel cleaning step after the crossover operation that zeros out certain genes with a probability $\lambda_{t0}$. This is important to reduce the perturbation visibility $\xi$ and ensure the attack remains imperceptible.

Notwithstanding, we realized a universal, non-conditional pixel cleaning operator will indefinitely decrease the perturbation magnitude, even past the $\epsilon$ constraint, as we have disabled elitism. This harms the misclassification score as disruptive pixels will continue to be zeroed out. Hence we turn off the pixel cleaning operator for all chromosomes in a generation that has reached a $\xi$ below that of the constraint.

\subsection{5. Elitism \& Selection}

The selection part of evolutionary algorithms is crucial in deciding which offsprings survive and which genes circulate. We began with a default configuration of keeping the best solution from previous generations. This seemingly innocuous positioning locked us with a large-norm solution, which never altered for the remaining generations. It turns out that a random configuration in the first generation, due to having the largest norm, never becomes subject to pixel cleaning or mutation operations. In other words, it always gets carried forward to the next generation, whereas the children are subject to the pixel cleaning step (often decreasing the misclassification rate). Its children, due to not improving as fast as the quickly-diminishing $\epsilon$ constraint, never supersede it as the best solution in a generation. 

To combat this, we disabled any type of forced elitism, making it such that the best solutions from previous generations need to re-qualify each round. We also do not keep any parents from previous generations to tighten the re-qualification. This frees the GA to accept new offspring with smaller norms once they produce a competitive misclassification score - significantly increasing the convergence speed. 

After this, we stayed faithful to a standard tournament approach for parent selection, maintaining a middle ground between diversity and exploration.

\subsection{5. Batch Loading}

To ensure true universality across the dataset, we periodically adjust the batch of images for fitness calculation. This ensures we are not overfitting to the same batch of images, causing the final perturbation to be overly specialized to only 64 images (we have a batch size of 64) and not the entire dataset. We also force the GA to find a perturbation that consistently fools multiple subsets, leading to a more universal attack. 

Initially, we tried loading a new batch of images each generation. This proved to have too short an evolutionary window and gave the algorithm chaotic signals. The best solution might do well on one batch but fail on the next, occasioning feedback which can be too noisy to lead to stable improvements. 

Hence, we arrived at a practical approach to change the batch every 4 generations. This way, we partially converge on each subset's vulnerability, before forcing the perturbation to adapt to the new subset in hopes of improving generalization. 

\section{Data Description}

We evaluate our universal adversarial perturbation approach on the \textbf{ImageNet-1K} dataset, a widely used benchmark for large-scale visual recognition \cite{imagenet_data}. This dataset, derived from the ImageNet project, contains 1,000 object categories with over 1.2 million training images and 50,000 validation images. Each image is typically preprocessed to a resolution of 224 x 224 pixels and normalized by channel-wise mean and standard deviation:

\begin{align*}
    \text{mean} &= [0.485, 0.456, 0.406]\\
    \text{std} &= [0.229, 0.225, 0.226]
\end{align*}

Our experiment primarily target \textbf{GoogLeNet}, \textbf{ResNet50}, and \textbf{ViT-B/16} models - networks trained or fine-tuned on ImageNet with top-tier performances. This ensures that the resulting universal perturbations are tested against a diverse set of architectural styles (inception-like, residual, and transformer-based) to ensure architecture universality. All images remain in float format $[0, 1]$ domain after normalization, consistent with standard PyTorch preprocessing. By testing on ImageNet-1K, we demonstrate our method's ability to scale to a complex dataset with high inter-class variability. 



\begin{figure*}[t]  
\centering
\includegraphics[width=1.0\textwidth]{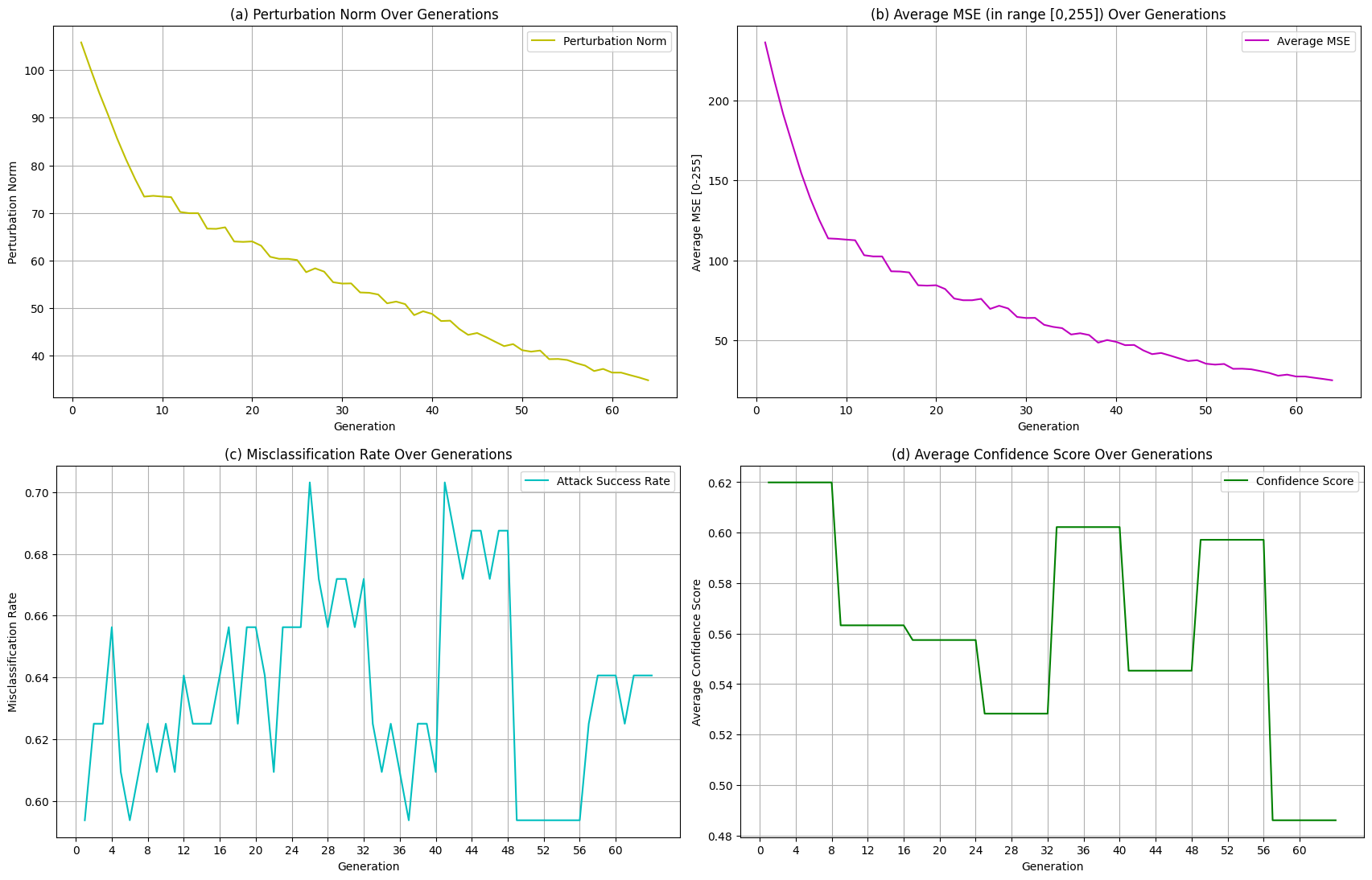}
\caption{(a) \textbf{Perturbation Norm}: L2 Norm of the universal adversarial perturbation over 64 generations (b) \textbf{Average Mean Squared Error (MSE)}: Average MSE between the original image and the attacked image (with perturbation superimposed on top). This is computed in the [0, 255] unnormalized image domain. (c) \textbf{Misclassification Rate}: Rate of misclassification on the batch of 64 images (with a new batch loaded every 4 generations) (d) \textbf{Average Confidence Score}: The average confidence score of the model on correct labels on each batch of 64 images.}\label{fig:metrics}
\end{figure*}

\section{Results} \label{sec:results}

This work demonstrates a float-coated, penalty-driven evolutionary framework for universal adversarial perturbations, offering a modern and Pytorch-compatible approach that seamlessly integrates into contemporary deep learning pipelines. By removing integer-encoded flips, we enable continuous, gradient-like updates to the perturbation, yielding a smoother optimization path. Our single-objective fitness function - combined with dynamic $\epsilon$ and the removal of elitism - not only fosters visually imperceptible perturbations but also ensures that smaller-norm offsprings can replace large-norm parents as soon as misclassification goals are met. 

As illustrated in Figure \ref{fig:metrics} (subplots (a)-(d)), our approach achieves a steady decrease in the perturbation norm (from $\sim110$ to $\sim40$), while the average MSE in the $[0,255]$ domain also significantly drops across only 64 generations. The misclassification rate remains stably high throughout, demonstrating robust attack performance even as the norm and MSE decline. Notably, the model's confidence score on the correct labels falls steadily towards the end, indicating that our perturbations also reduce the network's certainty in its (wrong) predictions. 

Gupta et al. \cite{eupa} were able to reduce the GoogLeNet from an accuracy of 0.698 to 0.446 with an attack having an average MSE of 393.59. This was achieved after 1408 epochs. We were able to deceive the network to create an accuracy of 0.406 after only 64 evolutionary epochs. Furthermore, our perturbation has an average MSE of less than 50, making it less perceptible yet more deadly at deceiving the network. 

Table \ref{tab:three_networks} also highlights the universality of the final perturbation across three diverse networks. For the three models of GoogLeNet, ResNet-50, and ViT-B/16, the attack has been to diminish the accuracy by 29.2\%, 45.2\%, and 22.9\%, respectively. While the perturbation is calculated only using GoogLeNet to reduce computational overhead, it turns out to be more effective on ResNet-50 with almost 4 times as many parameters. This is an interesting result and confirms our approach not only outperforms earlier evolutionary UAP methods in speed and norm reduction, but also maintains higher misclassification effectiveness across a plethora of network architectures.

\section{Discussion} \label{sec:discussion}

This iterative, multi-generation approach remains computationally intensive for large datasets of high-resolution images. We have tried to mitigate this by limiting population size to 50 and adopting partial sub-population mutation. Further work in optimizing the algorithm could likely yield promising results. 

Moreover, we primarily tested a small set of images in each batch, focusing on universal perturbations but not fully sampling all ImageNet classes in each generation. Having an aggregate fitness - where we compute misclassification for mini-batches and then average for the overall fitness - could be more robust at the cost of more computation per generation. This further subdues overfitting and is certainly a direction of interest. We could also have a two-phase approach, where for the first phase, focus on finding a decent universal solution with a single batch, and for the second phase, focus on refining the perturbation's generality by periodically switching or combining multiple batches. 

In this work, we also extend the universality of the perturbations by verifying on different architectures of GoogLeNet, ResNet, and Vision Transformers (ViT). Instead of verifying at the end, we could use ensemble fitness methods: aggregating fitness across the models during the evolutionary process to ensure true universality. Testing the method on even more advanced networks, regularization to control sparsity, along with multi-batch universal attacks, remain active directions and should be easily integrated by the flexible penalty approach.

\section{Conclusion}

In summary, our results demonstrate that a penalty-based single-objective GA with dynamic scheduling, conditional pixel cleaning, and float-coded genes can outperform existing Universal Perturbation Attack methods in terms of universalality across multiple networks, lower final perturbation norm, as well as higher misclassification success. This highlights the efficacy of adopting more modern design choices within an evolutionary search for universal adversarial attacks. 




\footnotesize
\bibliographystyle{apalike}
\bibliography{example} 

\end{document}